\documentclass[pdflatex,sn-mathphys-num]{sn-jnl}

\usepackage{graphicx}%
\usepackage{multirow}%
\usepackage{amsmath,amssymb,amsfonts}%
\usepackage{amsthm}%
\usepackage{mathrsfs}%
\usepackage[title]{appendix}%
\usepackage{xcolor}%
\usepackage{textcomp}%
\usepackage{manyfoot}%
\usepackage{booktabs}%
\usepackage{algorithm}%
\usepackage{algorithmicx}%
\usepackage{graphicx}
\usepackage{booktabs}
\usepackage{tabularx}
\usepackage{array}
\usepackage{hyperref}
\usepackage[capitalise,nameinlink,noabbrev]{cleveref}
\usepackage{algpseudocode}%
\usepackage{listings}%
\definecolor{tabfirst}{HTML}{F4B6B6}   
\definecolor{tabsecond}{HTML}{FBDDB8}  
\newcommand{\first}[1]{\cellcolor{tabfirst}\textbf{#1}}
\newcommand{\second}[1]{\cellcolor{tabsecond}#1}
\newcommand{\cmark}{\ding{51}}
\newcommand{\xmark}{\ding{55}}
\usepackage{graphicx}
\usepackage{booktabs}
\usepackage{multirow}
\usepackage{pifont}

\usepackage{booktabs}
\usepackage{multirow}
\usepackage{graphicx}
\usepackage[table]{xcolor}
\usepackage{pifont}

\providecommand{\cmark}{\ding{51}}
\providecommand{\xmark}{\ding{55}}

\theoremstyle{thmstyleone}%
\theoremstyle{thmstyletwo}%

\theoremstyle{thmstylethree}%

\begin{document}

\title[Article Title]{Every Pixel Is an Option: A System One Model for Medical Ultrasound Biometry}


\author*[1]{\fnm{Moein} \sur{Heidari}}
\email{moein.heidari@ubc.ca}

\author[1]{\fnm{Junbo} \sur{Rao}}
\author[1]{\fnm{Yudan} \sur{Chen}}
\author[1]{\fnm{Yasamin} \sur{Medghalchi}}
\author[1]{\fnm{River} \sur{Jiang}}
\author[2]{\fnm{Wenjin} \sur{Chen}}
\author[2]{\fnm{David J.} \sur{Foran}}
\author[1]{\fnm{Jasmine} \sur{Grewal}}
\author[1]{\fnm{Ilker} \sur{Hacihaliloglu}}


\affil*[1]{
\orgname{University of British Columbia},
\orgaddress{\city{Vancouver}, \country{Canada}}
}

\affil[2]{
\orgname{Rutgers Cancer Institute},
\orgaddress{\city{New Brunswick}, \country{USA}}
}

\abstract{
\textbf{Purpose:}
Ultrasound biometry is widely used across various clinical applications, including prenatal and cardiac examinations, to obtain quantitative measurements by placing calipers at anatomical landmarks. However, these measurements remain subject to interoperator variability as they rely largely on manual caliper placement. Conventional methods often use specialized prediction heads for different ultrasound tasks, whereas vision-language models offer flexibility through text instructions. However, direct coordinate prediction does not explicitly model spatial likelihood distributions, limiting the representation of localization ambiguity.
\textbf{Methods:}
We propose Kev-Bio, a vision-language model that uses clinical measurement questions to guide ultrasound landmark detection. Inspired by System One decision models, Kev-Bio uses decision representations of the language model to generate queries for each caliper endpoint. These queries are matched against shared dense ultrasound features to produce spatial probability maps, from which landmark coordinates and clinical measurements are derived. The model is trained end-to-end using both landmark localization and clinical measurement losses.
\textbf{Results:}
On the nine-view FU/GU Biometry 2026 dataset, Kev-Bio achieves a mean radial error (MRE) of 22.50 pixels and a clinical measurement mean absolute error (MAE) of 15.38, outperforming the compared baselines in overall performance.
\textbf{Conclusion:}
These results demonstrate the effectiveness of Kev-Bio for ultrasound landmark detection and clinical measurement across different anatomical views. Our code and models are available at \href{https://github.com/moeinheidari7829/Kev-Bio}{https://github.com/moeinheidari7829/Kev-Bio}.
}

\keywords{Ultrasound Biometry, Vision-language Models, Landmark Detection, Large Language Models}



\maketitle

\section{Introduction}

Ultrasound is widely used to examine anatomy during routine care and to guide clinical procedures. It provides real-time images without exposing patients to ionizing radiation \cite{ceriani2025point}. Beyond visual assessment, many clinical examinations also require quantitative measurements of anatomical structures. These measurements are obtained by placing calipers at predefined landmarks and calculating distances, angles, or circumferences. Choosing the correct points requires knowledge of both anatomy and the measurement protocol. Although standardized training can yield high reproducibility, measurements can still differ between observers \cite{donadono2022training}. Automatic caliper placement could reduce this manual effort and make quantitative assessment more consistent \cite{mor2023deep}.

Accurate anatomical landmark localization remains challenging in ultrasound imaging. Speckle noise, acoustic shadowing, and indistinct tissue boundaries can obscure anatomical structures, while changes in probe orientation alter their appearance. A single ultrasound image may support multiple clinical measurements, each requiring a distinct set of landmarks. Models developed for these tasks commonly use a shared image encoder with separate decoders or landmark-specific parameters~\cite{rasaeeunified}. Although these architectures can localize predefined landmarks, each measurement task remains tied to a fixed output head or task-specific parameter set, making adaptation to new anatomical structures or
measurement tasks challenging.

Previous studies have used multimodal language models to predict segmentation masks and point locations from text and image representations \cite{lai2024lisa,jhang2026enhancing}. Ultrasound biometry, however, requires precise localization of multiple measurement endpoints, whose anatomical roles vary across tasks. The central challenge is to identify which endpoints a clinical instruction refers to and locate each endpoint accurately in the image within a shared framework.

Recent System One models, such as Jev~\cite{deusser2026evaluating} and its open-weight counterpart Kev~\cite{palmer2026kev}, formulate prediction as scoring candidate responses rather than generating answers autoregressively. While primarily explored for decision-support tasks, this formulation offers potential for spatial localization by treating image locations as candidate responses to clinical localization questions.

Building on this insight, we introduce \emph{Kev-Bio}, a unified framework that translates clinical measurement questions into landmark queries and grounds them in dense ultrasound features to localize anatomical landmarks. Specifically, we initialize Qwen3.5 with Kev~\cite{team2026qwen3,palmer2026kev} to interpret
the ultrasound image and measurement instructions,
generating queries for the required anatomical landmarks. These queries interact with a shared dense feature map combining representations from Qwen3.5, USFM, and DINOv2~\cite{team2026qwen3,jiao2024usfm,oquab2023dinov2} to produce landmark probability maps. The maps provide spatial information about localization ambiguity, while joint landmark and measurement supervision guides training. We evaluate this formulation across nine ultrasound views through comparisons with established methods and controlled ablation studies.

\section{Related Work}

\paragraph{Landmark detection in medical images.}
Landmark detectors commonly predict coordinates or heatmaps at predefined anatomical points. nnLandmark provides a self-configuring approach to three-dimensional landmark detection, while MedSapiens adapts pose estimation representations to medical images \cite{ertl2026nnlandmark,elbatel2026medsapiens}. For ultrasound biometry, CAMS and IMPACT-L combine shared image encoders with decoders specific to each task \cite{licams,rasaeeunified}. Echo-CoPilot illustrates the use of learned measurement tools in a broader system for echocardiographic interpretation \cite{heidari2025echo}. Our method instead derives endpoint queries from the representation of a clinical measurement question.

\paragraph{Language models and spatial prediction.}
LISA uses a language model state to predict segmentation masks, while recent point grounding work predicts heatmaps from multimodal representations \cite{lai2024lisa,jhang2026enhancing}. We build on this connection between language and spatial prediction, but focus on groups of caliper points whose geometry defines a clinical measurement. Kev is an open decision model trained on Qwen3.5 \cite{palmer2026kev}. We initialize the Qwen3.5 language model with the pretrained Kev adapter and train a separate pixel pointer to localize ultrasound landmarks.

\section{Method}
\label{sec:method}
\begin{figure*}[t]
  \centering
  \includegraphics[width=\textwidth]{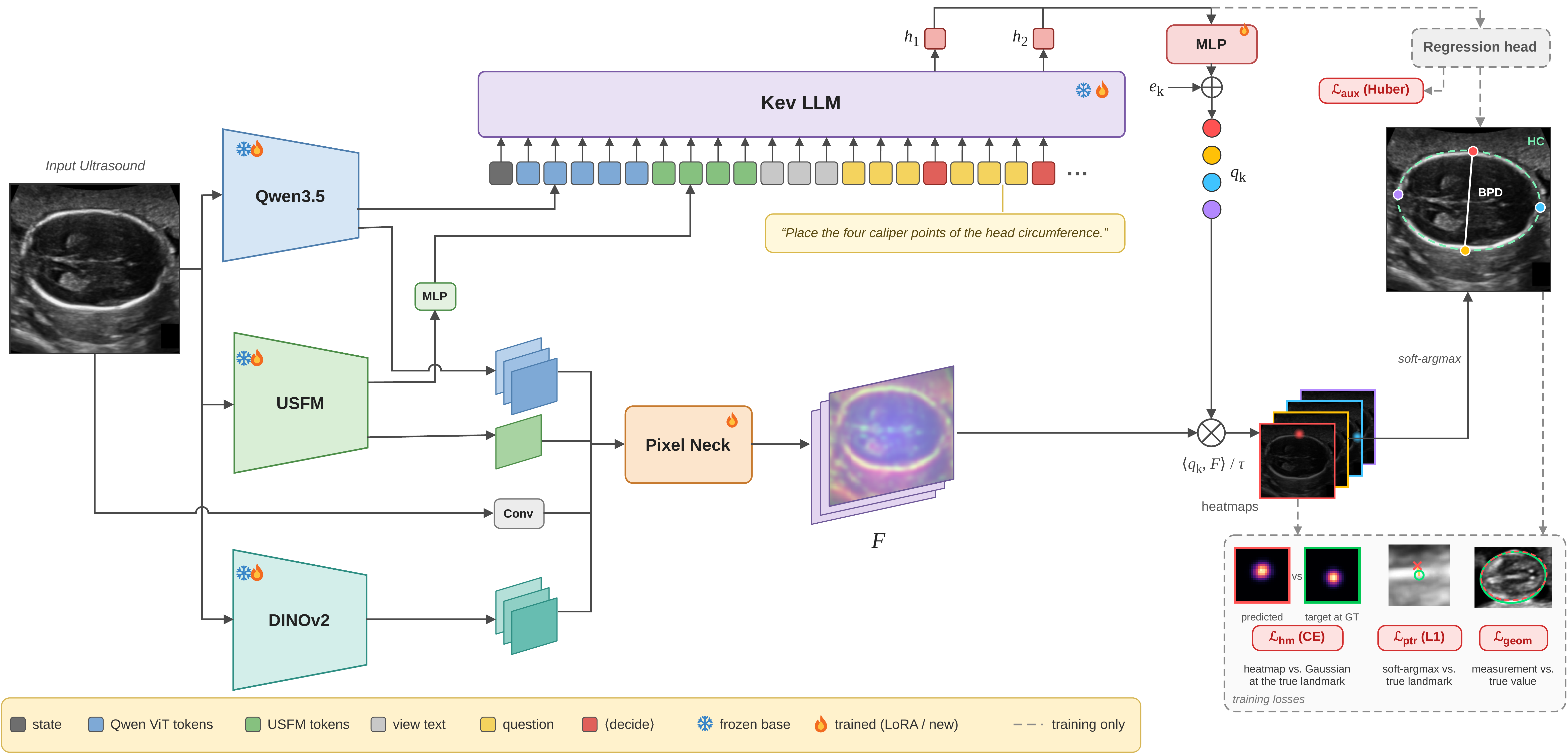}
  \caption{\textbf{Overview of Kev-Bio.}
Qwen3.5 and USFM provide visual tokens to the language model, while DINOv2 contributes additional localization features. Clinical questions are encoded into decision states at $\langle\mathrm{decide}\rangle$ tokens, which generate queries for each landmark endpoint. A convolutional-based Pixel Neck combines features from the three visual encoders into a shared dense feature map. Endpoint queries are matched against this map to produce spatial probability maps, from which landmark coordinates and clinical measurements are derived.}
  \label{fig:method}
\end{figure*}

Let $I$ be an ultrasound image of width $W$ and height $H$, with a known anatomical view $v$. Each view has a predefined set of clinical measurements $\mathcal{M}(v)$, where each measurement $m \in \mathcal{M}(v)$ is associated with a clinical question $t_m$ and the number of required anatomical landmarks is $K_m \in \{2,3,4\}$. Kev-Bio predicts these landmarks and computes the corresponding measurements from their coordinates. As illustrated in \cref{fig:method}, the model combines a multimodal language model with a shared dense visual feature map. The language model produces a decision representation for each measurement, which is then transformed into endpoint-specific queries for locating landmarks in the feature map.

\subsection{Instruction-guided Multimodal Encoding}
\label{sec:clinical_encoding}

We use the Qwen3.5 vision encoder \cite{team2026qwen3} and an ultrasound foundation model (USFM) \cite{jiao2024usfm} to extract visual features from the input image. Their visual tokens are denoted by $Z^Q(I)$ and $Z^U(I)$, respectively. The USFM tokens are projected into the language model's embedding space using a learned projection network $\mathcal{P}_U$. The input also includes a view instruction $t_v$ describing the anatomical view and clinical domain, followed by a clinical measurement question $t_m$ for each measurement $m \in \mathcal{M}(v)$. Each question starts with a \texttt{\textless q\textgreater} token and ends with a \texttt{\textless decide\textgreater} token. The input sequence is
\begin{equation}
X = [\langle\mathrm{state}\rangle,\,Z^Q(I),\,\mathcal{P}_U(Z^U(I)),\,t_v,\,
(\langle\mathrm{q}\rangle,t_m,\langle\mathrm{decide}\rangle)_{m\in\mathcal M(v)}].
\label{eq:input_sequence}
\end{equation}
The measurement questions are concatenated in a predefined order.
We initialize the Qwen3.5 language backbone (\textbf{LM}) with the pretrained Kev adapter. The sequence $X$ is processed in a single forward pass, and the hidden state at each $\langle\mathrm{decide}\rangle$ token is extracted as the corresponding measurement representation, $\mathbf h_m = \mathrm{LM}(X)_{\langle\mathrm{decide}\rangle_m}$.

\subsection{Dense Ultrasound Feature Fusion}
\label{sec:dense_representation}

\noindent We combine visual representations from Qwen3.5
\cite{team2026qwen3}, USFM \cite{jiao2024usfm}, and
DINOv2 \cite{oquab2023dinov2} to construct a shared dense
feature map for anatomical landmark localization.
These encoders provide complementary language-aligned,
ultrasound-specific, and multilevel spatial features,
respectively, allowing the fused representation to capture
both anatomical context and fine spatial details.

\noindent These features
are aligned to a common spatial resolution using bilinear
interpolation, normalized, and concatenated. A dense fusion module, termed the \textbf{Pixel Neck}, projects and progressively upsamples the combined features while incorporating shallow convolutional features from the input image to preserve fine spatial details:

\begin{equation}
\mathbf F=\operatorname{norm}\!\left(
\mathcal N\bigl(
\mathcal A(Z^Q_{\mathrm{dense}},
Z^U_{\mathrm{dense}},Z^D_{\mathrm{dense}}),
C(I)
\bigr)\right)
\in\mathbb R^{d_p\times H_p\times W_p}.
\label{eq:pixel_map}
\end{equation}
Here, $\mathcal A$ denotes spatial alignment and
concatenation, $\mathcal N$ represents the dense fusion
module, Pixel Neck, and $C(I)$ denotes shallow image features.
The resulting map, F, is L2-normalized along the channel
dimension and shared across measurement tasks.

\subsection{Language Conditioned Pixel Pointer}
\label{sec:pixel_pointer}

The pixel pointer localizes anatomical landmarks by matching endpoint queries against a shared dense feature map to produce spatial probability maps.
The decision representation $\mathbf h_m$ is passed through an MLP to generate endpoint queries, with a learnable endpoint embedding $\mathbf e_{m,k}$ added to distinguish the $k$th landmark of measurement $m$. The resulting queries are L2-normalized:
\begin{equation}
\mathbf q_{m,k}=\operatorname{norm}\!\left(
\operatorname{reshape}\bigl(\operatorname{MLP}(\mathbf h_m)\bigr)_k+\mathbf e_{m,k}\right),
\qquad k=1,\ldots,K_m.
\label{eq:queries}
\end{equation}

Then, the similarity between $\mathbf q_{m,k}$ and $\mathbf F(i,j)$ is calculated at each spatial location using a dot product, followed by a spatial softmax to obtain the landmark probability maps:
\begin{equation}
S_{m,k}(i,j)=\frac{\mathbf q_{m,k}^{\top}\mathbf F(i,j)}{\tau},
\qquad
P_{m,k}(i,j)=\operatorname{softmax}_{i,j}\bigl( S_{m,k}(i,j)\bigr),
\label{eq:pointer}
\end{equation}
where $\tau>0$ is a learnable temperature parameter. The probability map $P_{m,k}$ represents the spatial distribution of the $k$th landmark for measurement $m$.

Landmark coordinates are calculated using differentiable soft-argmax:
\begin{equation}
\hat{\mathbf p}_{m,k}=(W,H)\odot\sum_{i,j}P_{m,k}(i,j)\mathbf c_{i,j},
\qquad
\mathbf c_{i,j}=\left(\frac{j+1/2}{W_p},\frac{i+1/2}{H_p}\right).
\label{eq:softargmax}
\end{equation}
Here, $\mathbf c_{i,j}$ denotes the normalized center of each spatial cell, and $\odot$ represents elementwise multiplication.

\subsection{Heatmap and Measurement Supervision}
\label{sec:geometry_supervision}

The predicted landmarks for measurement \(m\) are passed to a measurement-specific geometric function to calculate a biometric value,
$\hat y_m=g_m(\hat{\mathbf p}_{m,1},\ldots,\hat{\mathbf p}_{m,K_m})$.
The geometric function computes a distance, an angle, or a circumference approximation from the corresponding calipers depending on the measurement. 

\noindent Training jointly supervises the predicted landmark coordinates, their spatial probability maps, and the clinical measurement values calculated from these coordinates. The pointer loss $\mathcal L_{\mathrm{ptr}}$ calculates the L1 error between predicted and ground-truth landmark coordinates and normalizes it by the image diagonal length. The heatmap loss $\mathcal L_{\mathrm{hm}}$ calculates the cross-entropy between the predicted spatial probability map and a normalized Gaussian distribution centered at the ground-truth landmark. This transforms point-based
landmark supervision into a continuous probability field,
encouraging high confidence near the true location while
allowing spatial uncertainty around it. The geometric loss $\mathcal L_{\mathrm{geom}}$ calculates the absolute error between the predicted and ground-truth measurement values and normalizes it by the training-set median of the corresponding measurement. An auxiliary regression head predicts endpoint coordinates and a measurement value directly from each decision state and uses Huber loss $\mathcal L_{\mathrm{aux}}$ to supervise both outputs. The complete objective is
\begin{equation}
\mathcal L=\mathcal L_{\mathrm{ptr}}
+\lambda_{\mathrm{hm}}\mathcal L_{\mathrm{hm}}
+\lambda_{\mathrm{geom}}\mathcal L_{\mathrm{geom}}
+\lambda_{\mathrm{aux}}\mathcal L_{\mathrm{aux}}.
\label{eq:objective}
\end{equation}
The auxiliary regression head provides additional supervision during training. During inference, the landmark coordinates and the corresponding measurement values are obtained from the pixel pointer and the geometric functions, respectively.

\section{Results}\label{Results}

\subsection{Dataset and Implementation Details}
We evaluate Kev-Bio on the FU/GU Biometry 2026 benchmark, introduced at the MICCAI 2026 FoundUS challenge~\cite{bai_2026_19736827}. The dataset spans nine ultrasound views across three clinical domains: cardiac, prenatal, and intrapartum imaging (\cref{tab:dataset}). The labelled benchmark contains 6,743 images, with 60 view-specific landmark definitions and 29 clinical measurements. These measurements include distances, ellipse circumferences, and angles, with each view requiring between one and eleven measurements. The dataset is highly imbalanced, ranging from 38 labelled images for IVC to 4,000 for AoP. An additional 191,170 unlabelled frames are provided but are not used in our experiments. Since the official challenge test annotations are unavailable for offline evaluation, we perform five-fold cross-validation on the labelled data, with approximately 1,350 images held out per fold. All methods are trained and evaluated using the same partitions.

\begin{table}[t]
\caption{FU/GU Biometry 2026 labelled dataset. Samples: number of labelled images; LM: landmarks per image; Meas.: clinical measurements per image (D: distance, C: ellipse circumference, A: angle).}
\label{tab:dataset}
\centering

\begin{tabularx}{\linewidth}{
@{}
l
>{\raggedright\arraybackslash}X
r
r
l
@{}
}
\toprule
Task & View & Samples & LM & Meas. \\
\midrule

\multicolumn{5}{@{}l}{\emph{Cardiac}} \\
A4C   & Apical four-chamber       & 108   & 16 & 8 D \\
PLAX  & Parasternal long-axis     & 87    & 22 & 11 D \\
PSAX  & Parasternal short-axis    & 49    & 4  & 2 D \\
IVC   & Subcostal inferior vena cava & 38 & 2 & 1 D \\

\midrule
\multicolumn{5}{@{}l}{\emph{Prenatal}} \\
HC    & Fetal head (transthalamic) & 999 & 4 & 1 D + 1 C \\
FA    & Fetal abdomen             & 500   & 4 & 1 C \\
Femur & Fetal femur                & 702   & 2 & 1 D \\
FUGC  & Transvaginal cervix        & 260   & 2 & 1 D \\

\midrule
\multicolumn{5}{@{}l}{\emph{Intrapartum}} \\
AoP   & Transperineal (angle of progression)
      & 4,000 & 4 & 1 D + 1 A \\

\midrule
\textbf{Total} & \textbf{9 views}
& \textbf{6,743} & \textbf{60} & \textbf{29} \\

\bottomrule
\end{tabularx}
\end{table}

\noindent We initialize the language backbone from Kev~\cite{palmer2026kev}, based on Qwen3.5-0.8B, and fine-tune it with  LoRA adapters (r=16). The Qwen vision encoder and USFM are adapted with rank-16 LoRA, while DINOv2 ViT-L/14 uses rank-32. We set $\lambda_{\mathrm{hm}}=0.1$, $\lambda_{\mathrm{geom}}=0.2$, and $\lambda_{\mathrm{aux}}=1$ and train for 24 epochs using AdamW with a weight decay of 0.05 and an effective batch size of 8. Learning rates are $10^{-3}$ for newly initialized modules, $10^{-4}$ for the language model and DINOv2 adapters, and $5\times10^{-5}$ for the remaining adapters. For all methods and baselines, we adopt the publicly released codebases and corresponding pretrained model weights. Training samples are drawn uniformly across the nine views, with photometric and affine augmentations following~\cite{panagiotakopoulouself}. All experiments were conducted on NVIDIA H100 GPUs (80 GB GPU memory).


\subsection{Comparison with Prior Methods}

We compare Kev-Bio with two general-purpose landmark detectors, nnLandmark~\cite{ertl2026nnlandmark} and MedSapiens~\cite{elbatel2026medsapiens}, and four methods developed for the FU/GU Biometry challenge: CAMS~\cite{licams}, IMPACT-L~\cite{rasaeeunified}, FMUB~\cite{wuunified}, and US-Foundation~\cite{panagiotakopoulouself}. We report mean radial error (MRE) for landmark localization and mean absolute error (MAE) for clinical measurements, both in pixels (degrees for AoP) averaged across the nine views.

\noindent As shown in \cref{tab:main}, Kev-Bio achieves the lowest average error on both metrics among the reported methods, with 22.50 pixels MRE and 15.38 MAE, compared with 23.88 and 17.47 for the strongest respective baselines. It achieves the best landmark localization on eight of the nine views and the lowest measurement error on six, spanning both cardiac and prenatal imaging, with a comparable parameter count. The improvement is particularly notable in fetal head biometry, where MRE decreases from 17.03 to 13.76 pixels. Notably, these results are obtained without using the additional unlabelled challenge data leveraged by CAMS and FMUB. Although competing methods achieve slightly lower measurement errors on femur length, cervical length, and angle of progression, Kev-Bio maintains the lowest average error across both metrics, demonstrating its effectiveness across diverse anatomical views and measurement tasks.

\begin{table*}[t]
\centering
\caption{Comparison on 5-fold cross-validation of the FU/GU Biometry 2026 training set (lower is better). Top: mean radial error of landmarks (MRE, px). Bottom: mean absolute error of clinical measurements (MAE; px, degrees for AoP). Per-task values are averaged over five folds; Avg.\ is the macro mean over nine tasks $\pm$ s.d.\ over folds. Params: total trainable parameters in millions. \colorbox{tabfirst}{\textbf{Best}} and \colorbox{tabsecond}{second best} per column among reported values.}
\label{tab:main}
\setlength{\tabcolsep}{3pt}
\resizebox{\textwidth}{!}{%
\begin{tabular}{lccccccccccc}
\toprule
\multirow{2}{*}{Method}
& \multirow{2}{*}{Params (M)}
& \multicolumn{4}{c}{Cardiac}
& \multicolumn{4}{c}{Prenatal}
& Intrapartum
& \multirow{2}{*}{Avg.} \\
\cmidrule(lr){3-6}
\cmidrule(lr){7-10}
\cmidrule(lr){11-11}
& & A4C & PLAX & PSAX & IVC
& HC & FA & Femur & FUGC & AoP & \\
\midrule

\multicolumn{12}{l}{\emph{MRE (px) $\downarrow$}} \\
\midrule
\multicolumn{12}{l}{\textit{General Baselines}} \\

nnLandmark~\cite{ertl2026nnlandmark}
& 140.1 & 44.23 & 26.71 & 58.87 & 45.92
& 25.45 & 19.36 & 22.28 & 6.99
& 8.79 & 28.73 $\pm$ 4.12 \\

MedSapiens~\cite{elbatel2026medsapiens}
& 24.0 & 118.85 & 76.49 & 108.41 & 130.09 & 35.32 & 24.34 & 24.75 & 11.56 & 9.61 & 59.94 $\pm$ 4.15 \\

\midrule
\multicolumn{12}{l}{\textit{Challenge Baselines}} \\

CAMS~\cite{licams}
& 35.5 & 44.92 & \second{20.99} & \second{42.76}
& \second{36.31} & 18.07 & 17.04
& \second{19.29} & 6.86 & 8.71
& \second{23.88} $\pm$ 2.10 \\

IMPACT-L~\cite{rasaeeunified}
& 144.4 & 44.40 & 23.32 & 60.22 & 51.72
& \second{17.03} & \second{15.95}
& 19.66 & 7.54 & 9.41
& 27.70 $\pm$ 3.18 \\

FMUB~\cite{wuunified}
& 29.5 & \second{41.73} & 21.45 & 48.80 & 41.17
& 18.58 & 16.58 & 20.39 & 6.90 & 9.22
& 24.98 $\pm$ 3.41 \\

US-Foundation~\cite{panagiotakopoulouself}
& 56.1 & 44.69 & 22.15 & 51.33 & 45.10
& 17.70 & 16.48 & 20.16
& \second{6.61} & \first{8.44}
& 25.85 $\pm$ 2.87 \\

\midrule
Kev-Bio (ours)
& 35.0
& \first{40.88} & \first{20.95}
& \first{42.24} & \first{35.08}
& \first{13.76} & \first{15.86}
& \first{18.77} & \first{6.47}
& \second{8.48}
& \first{22.50 $\pm$ 2.10} \\

\midrule

\multicolumn{12}{l}{\emph{MAE $\downarrow$}} \\
\midrule
\multicolumn{12}{l}{\textit{General Baselines}} \\

nnLandmark~\cite{ertl2026nnlandmark}
& & 25.65 & 12.31 & 35.31 & 22.43
& 23.47 & 53.96 & 18.85 & 4.46
& 4.85 & 22.37 $\pm$ 1.47 \\

MedSapiens~\cite{elbatel2026medsapiens}
& & 59.77 & 40.27 & 75.64 & 105.43 & 27.16 & 71.68 & 19.87 & 7.50 & 5.34 & 45.85 $\pm$ 4.15 \\

\midrule
\multicolumn{12}{l}{\textit{Challenge Baselines}} \\

CAMS~\cite{licams}
& & 24.95 & \second{8.52}
& \second{22.45} & \second{9.90}
& 24.09 & 52.05 & \first{15.63}
& 4.32 & 5.33 & 18.58 $\pm$ 1.28 \\

IMPACT-L~\cite{rasaeeunified}
& & 18.82 & 9.37 & 36.08 & 11.18
& 15.80 & \second{50.94}
& \second{16.49} & 5.38 & 5.70
& 18.86 $\pm$ 1.20 \\

FMUB~\cite{wuunified}
& & \second{18.00} & 8.74
& 25.10 & 11.12 & 15.00
& 52.23 & 17.45 & 4.22 & 5.34
& \second{17.47} $\pm$ 0.90 \\

US-Foundation~\cite{panagiotakopoulouself}
& & 22.42 & 10.54 & 28.77 & 15.87
& \second{14.93} & 51.72 & 17.94
& \first{3.92} & \first{4.63}
& 18.97 $\pm$ 1.41 \\

\midrule
Kev-Bio (ours)
& & \first{16.12} & \first{8.42}
& \first{19.23} & \first{9.48}
& \first{10.59} & \first{49.22}
& 16.53 & \second{4.00}
& \second{4.84}
& \first{15.38 $\pm$ 1.24} \\

\bottomrule
\end{tabular}}
\end{table*}


\subsection{Qualitative Results}
\label{sec:quantitative_results}

\begin{figure*}[h]
  \centering
  \includegraphics[width=0.9\textwidth]{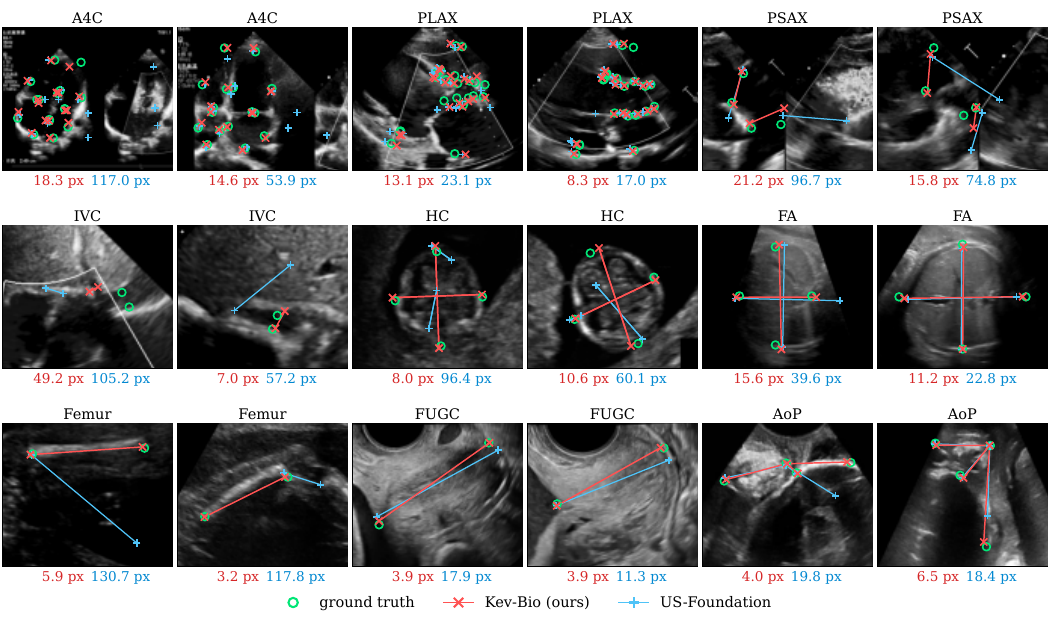}
  \caption{\textbf{Qualitative comparison across nine ultrasound views.} Landmark predictions from Kev-Bio (red) and US-Foundation (blue), with ground truth shown in green. Values indicate mean landmark error in pixels.}
  \label{fig:qualitative}
\end{figure*}

\cref{fig:qualitative} presents qualitative comparisons between Kev-Bio and US-Foundation~\cite{panagiotakopoulouself} across all nine ultrasound views. Kev-Bio places most landmarks closer to their annotated anatomical targets, whereas US-Foundation shows larger localization errors, including predictions on neighboring structures. This is particularly evident in the IVC, fetal head, and femur examples, where substantial localization errors are observed for US-Foundation despite accurate predictions from Kev-Bio. In the A4C example containing adjacent cardiac views, Kev-Bio correctly identifies the target anatomy, while US-Foundation places several landmarks on the neighboring view. These examples illustrate the ability of text-conditioned spatial pointing to preserve anatomical correspondence across diverse ultrasound measurements.

\begin{figure*}[h]
  \centering
  \includegraphics[width=0.9\textwidth]{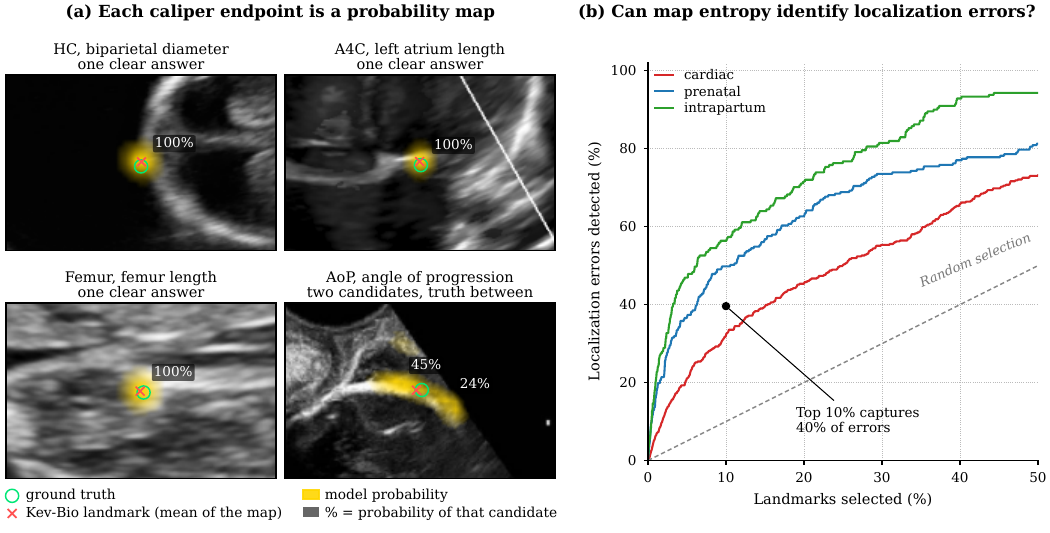}
  \caption{\textbf{Spatial probability maps and error detection.}
    (a) Landmark probability maps showing confident and ambiguous predictions.
    (b) Landmarks are ranked by decreasing map entropy, and curves show the fraction of localization errors captured in each clinical domain. Errors are defined using ground truth as deviations exceeding 5\% of the image diagonal. The black marker indicates the overall result, and the dashed line represents random selection.}
\label{fig:pixels_as_options}
\end{figure*}

\noindent Beyond landmark localization, Kev-Bio produces a spatial probability map for each endpoint (\cref{fig:pixels_as_options}a), whose entropy serves as an uncertainty score. To assess whether this score identifies localization failures, we consider predictions more than 5\% of the image diagonal from ground truth as errors. We then rank predictions by decreasing entropy and measure the fraction of errors captured among the highest-ranked predictions (\cref{fig:pixels_as_options}b). Inspecting the 10\% most uncertain predictions captures 40\% of errors, compared with 10\% under random selection. Error detection AUROCs reach 0.70, 0.76, and 0.86 for cardiac, prenatal, and intrapartum imaging, respectively. These results suggest that Kev-Bio's probability maps can help clinicians identify caliper placements that warrant closer inspection, enabling more targeted verification of automated measurements.

\begin{figure*}[!h]
  \centering
  \includegraphics[width=0.8\textwidth]{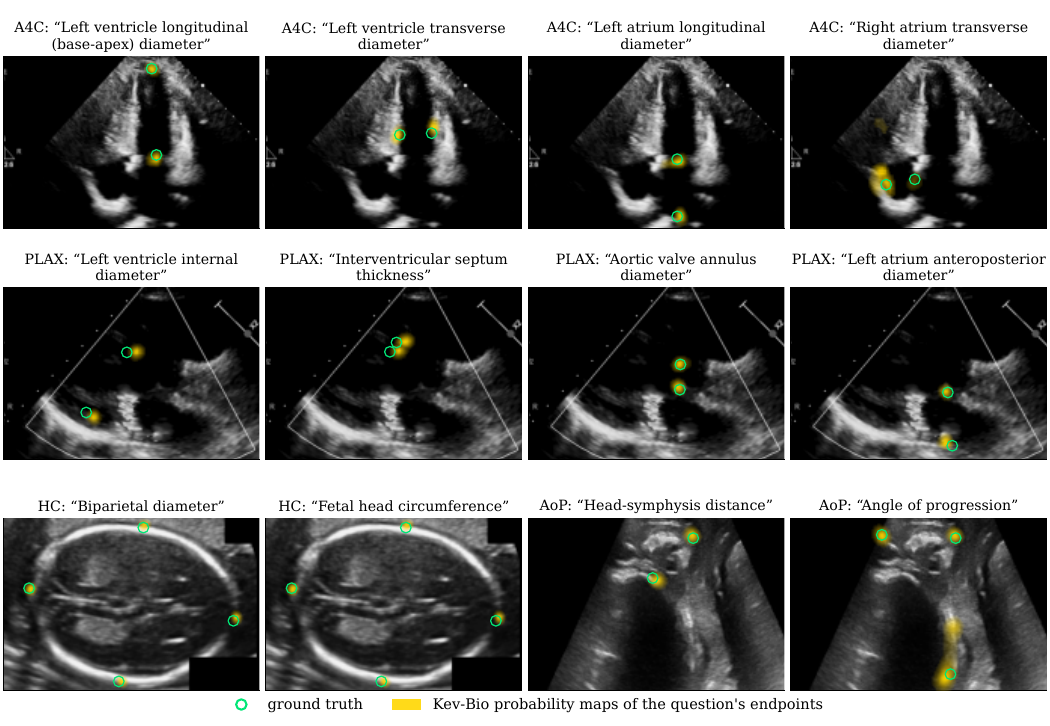}
\caption{\textbf{Question-conditioned landmark localization.} Different clinical measurement questions produce distinct landmark probability maps on the same ultrasound image. Green circles denote ground truth, and yellow regions indicate predicted spatial distributions.}

\label{fig:questions}

\end{figure*}

\noindent Finally, \cref{fig:questions} visualizes how clinical questions guide spatial localization. Given the same ultrasound image, different measurement questions produce distinct probability maps that concentrate on the corresponding anatomical landmarks. In A4C, for example, queries for ventricular length and atrial width highlight different anatomical endpoints, while related fetal head measurements yield consistent responses at shared landmarks. These observations illustrate how Kev-Bio adapts its spatial predictions to different measurement definitions through a shared, question-conditioned pixel pointer.

\subsection{Ablation Study}

We assess the contribution of each component by retraining model variants under identical settings, as reported in \cref{tab:ablation}. Removing the pixel pointer (replacing spatial localization with direct coordinate regression from the decision states) leads to the largest performance degradation, increasing MRE from 22.50 to 35.50 pixels and MAE from 15.38 to 23.64. This highlights the importance of matching measurement-specific queries to dense spatial features rather than directly regressing coordinates from the decision states. Removing DINOv2 also reduces performance, yielding 24.85 MRE and 16.94 MAE, indicating the benefit of complementary visual representations for precise landmark localization. Excluding the language model (LLM) and replacing its decision states with learned, image-conditioned measurement queries results in a smaller but consistent increase in error (23.66 MRE and 16.77 MAE), suggesting that language-conditioned queries contribute to localization accuracy even when measurement definitions are fixed.

\begin{table}[h]
\caption{Component ablation of Kev-Bio. We assess the impact of DINOv2, LLM-guided queries, and pointer-based localization. Results are the mean $\pm$ s.d.\ over five folds. Lower is better.}
\label{tab:ablation}
\begin{tabular*}{\textwidth}{@{\extracolsep\fill}lccccc}
\toprule
Variant & DINOv2 & LLM & Pointer & MRE (px) $\downarrow$ & MAE $\downarrow$ \\
\midrule
w/o DINOv2        & \xmark & \cmark & \cmark & 24.85 $\pm$ 2.66 & 16.94 $\pm$ 0.94 \\
w/o LLM           & \cmark & \xmark & \cmark & 23.66 $\pm$ 2.63 & 16.77 $\pm$ 1.23 \\
w/o pixel pointer & \cmark & \cmark & \xmark & 35.50 $\pm$ 3.12 & 23.64 $\pm$ 1.60 \\
\midrule
Kev-Bio (full)    & \cmark & \cmark & \cmark & \textbf{22.50 $\pm$ 2.10} & \textbf{15.38 $\pm$ 1.24} \\
\bottomrule
\end{tabular*}
\end{table}


\section{Discussion and Limitations}
Kev-Bio offers a unified approach to ultrasound biometry by handling different clinical measurements through a shared localization framework, as opposed to task-specific prediction heads. Beyond predicting landmark coordinates, its spatial probability maps provide information about ambiguous predictions, potentially helping clinicians identify caliper placements that require verification. Nevertheless, our evaluation is limited to predefined measurements on a single benchmark, despite spanning multiple anatomical regions. Generalization to unseen measurement instructions, anatomical structures, and different imaging conditions remains to be explored. Future work will investigate these directions, alongside uncertainty calibration and computational efficiency.

\section{Conclusion}
We introduced Kev-Bio, a System One framework that formulates medical ultrasound biometry as spatial decision-making. By grounding clinical measurement questions in dense visual features, the model localizes anatomical landmarks and derives clinical measurements within a unified framework. Experiments across nine ultrasound views demonstrate improvements over existing methods, while the predicted probability maps provide additional information about localization ambiguity.

\noindent
\subsubsection*{Acknowledgments.}
This work was supported by the Canadian Foundation for Innovation-John R. Evans Leaders Fund (CFI-JELF) program grant number 42816. Mitacs Accelerate program grant number AWD024298-IT33280. We also acknowledge the support of the Natural Sciences and Engineering Research Council of Canada (NSERC), [RGPIN-2023-03575]. Cette recherche a été financée par le Conseil de recherches en sciences naturelles et en génie du Canada (CRSNG), [RGPIN-2023-03575].

\noindent
\subsubsection*{Disclosure of Interests.} The authors have no competing interests in the paper.

\newpage
\bibliography{ref}

@inproceedings{panagiotakopoulouself,
  title     = {Self-Supervised Domain Adaptation with DINOv2-HRNet for Multi-Task Ultrasound Landmark Detection},
  author    = {Panagiotakopoulou, Anna and Barmperis, Alexandros and Katsigiannis, Vasileios E. and Matsopoulos, George K.},
  booktitle = {Medical Image Computing and Computer Assisted Intervention -- MICCAI 2026 Workshops and Challenges},
  year      = {2026},
  publisher = {Springer Nature Switzerland}
}

@article{ceriani2025point,
  title={Point of care ultrasound: focus on evidence for a critical appraisal},
  author={Ceriani, E and Schiavon, R and La Cava, L and Ruscitti, C and Cogliati, C},
  journal={European Journal of Internal Medicine},
  year={2025},
  publisher={Elsevier}
}

@article{donadono2022training,
  title={Training in obstetric ultrasound biometry: results from a multicenter reproducibility study},
  author={Donadono, Vera and Ambroise Grandjean, Ga{\"e}lle and Stegen, Marie-Louise and Collin, Adeline and Bertholdt, Charline and Casagrandi, Davide and Morel, Olivier and Napolitano, Raffaele},
  journal={Journal of Ultrasound in Medicine},
  volume={41},
  number={11},
  pages={2819--2825},
  year={2022},
  publisher={Wiley Online Library}
}

@inproceedings{rasaeeunified,
  title     = {Unified Coarse-to-Fine Landmark Localization for Multi-Domain Ultrasound Biometry},
  author    = {Rasaee, Hamza and Rivaz, Hassan},
  booktitle = {Medical Image Computing and Computer Assisted Intervention -- MICCAI 2026 Workshops and Challenges},
  year      = {2026},
  publisher = {Springer Nature Switzerland}
}

@inproceedings{lai2024lisa,
  title={Lisa: Reasoning segmentation via large language model},
  author={Lai, Xin and Tian, Zhuotao and Chen, Yukang and Li, Yanwei and Yuan, Yuhui and Liu, Shu and Jia, Jiaya},
  booktitle={2024 IEEE/CVF Conference on Computer Vision and Pattern Recognition (CVPR)},
  pages={9579--9589},
  year={2024},
  organization={IEEE}
}

@misc{team2026qwen3,
  title  = {{Qwen3.5}: Towards Native Multimodal Agents},
  author = {{Qwen Team}},
  year   = {2026},
  url    = {https://qwen.ai/blog?id=qwen3.5},
  note   = {Official Qwen3.5 release}
}

@misc{palmer2026kev,
  title        = {Kev-0.8B: a calibrated decision model on Qwen3.5-0.8B},
  author       = {Palmer, Jared},
  year         = {2026},
  howpublished = {\url{https://huggingface.co/jaredpalmer/kev-0.8b}},
  note         = {Kev 1.0}
}

@article{jiao2024usfm,
  title={Usfm: A universal ultrasound foundation model generalized to tasks and organs towards label efficient image analysis},
  author={Jiao, Jing and Zhou, Jin and Li, Xiaokang and Xia, Menghua and Huang, Yi and Huang, Lihong and Wang, Na and Zhang, Xiaofan and Zhou, Shichong and Wang, Yuanyuan and others},
  journal={Medical image analysis},
  volume={96},
  pages={103202},
  year={2024},
  publisher={Elsevier}
}

@article{oquab2023dinov2,
  title={Dinov2: Learning robust visual features without supervision},
  author={Oquab, Maxime and Darcet, Timoth{\'e}e and Moutakanni, Th{\'e}o and Vo, Huy and Szafraniec, Marc and Khalidov, Vasil and Fernandez, Pierre and Haziza, Daniel and Massa, Francisco and El-Nouby, Alaaeldin and others},
  journal={arXiv preprint arXiv:2304.07193},
  year={2023}
}

@article{ertl2026nnlandmark,
  title={nnLandmark: A Self-Configuring Method for 3D Medical Landmark Detection},
  author={Ertl, Alexandra and Denner, Stefan and Peretzke, Robin and Xiao, Shuhan and Zimmerer, David and Fischer, Maximilian and Bujotzek, Markus and Yang, Xin and Neher, Peter and Isensee, Fabian and others},
  journal={Proceedings of Machine Learning Research},
  volume={315},
  pages={894--927},
  year={2026}
}

@article{elbatel2026medsapiens,
  title={MedSapiens: Taking a pose to rethink medical imaging landmark detection},
  author={Elbatel, Marawan and Wang, Anbang and Liu, Keyuan and Mouheb, Kaouther and Almar-Munoz, Enrique and Lin, Lizhuo and Yang, Yanqi and Lekadir, Karim and Li, Xiaomeng},
  journal={Medical Image Analysis},
  pages={104015},
  year={2026},
  publisher={Elsevier}
}

@inproceedings{licams,
  title     = {CAMS: Coordinate-Aware Multi-task Semi-supervised Learning for Ultrasound Landmark Localization},
  author    = {Li, Chenke and Li, Xiaoshuang and Wang, Yuanhan and Wu, Beining and Qin, Feiwei and Chen, Yifei},
  booktitle = {Medical Image Computing and Computer Assisted Intervention -- MICCAI 2026 Workshops and Challenges},
  year      = {2026},
  publisher = {Springer Nature Switzerland}
}

@article{heidari2025echo,
  title={Echo-CoPilot: A Multiple-Perspective Agentic Framework for Reliable Echocardiography Interpretation},
  author={Heidari, Moein and Mehrabian, Ali and Roohi, Mohammad Amin and Jiang, River and Chen, Wenjin and Foran, David J and Grewal, Jasmine and Hacihaliloglu, Ilker},
  journal={arXiv preprint arXiv:2512.09944},
  year={2025}
}

@article{jhang2026enhancing,
  title={Enhancing Part-Level Point Grounding for Any Open-Source MLLMs},
  author={Jhang, Jin-Cheng and Wang, Fu-En and Yang, Xin and Qiao, Nan and Xia, Lu and Sun, Min and Kuo, Cheng-Hao},
  journal={arXiv preprint arXiv:2606.29267},
  year={2026}
}

@inproceedings{wuunified,
  title     = {A Unified Foundation-Model Framework with Semi-Supervised Consistency and Heatmap Ensembling for Nine-Task Ultrasound Biometry Regression},
  author    = {Wu, Yongji and Zhang, Jiansong and Yang, Ye and Yang, Shilong and Shen, Linlin},
  booktitle = {Medical Image Computing and Computer Assisted Intervention -- MICCAI 2026 Workshops and Challenges},
  year      = {2026},
  publisher = {Springer Nature Switzerland}
}

@misc{bai_2026_19736827,
  author       = {Bai, Jieyun and
                  Yaqub, Mohammad and
                  Ma, Jun and
                  Lekadir, Karim and
                  Gan, Jie and
                  Cai, Weidong and
                  Ni, Dong and
                  Li, Shuo},
  title        = {Foundation Model Challenge for Ultrasound Biometry},
  month        = apr,
  year         = 2026,
  publisher    = {Zenodo},
  doi          = {10.5281/zenodo.19736827},
  url          = {https://doi.org/10.5281/zenodo.19736827},
}

@article{deusser2026evaluating,
  title={Evaluating and benchmarking the System One model Jev},
  author={Deu{\ss}er, Tobias and Sparrenberg, Lorenz and Sifa, Rafet},
  journal={arXiv preprint arXiv:2609.37647},
  year={2026}
}

@article{mor2023deep,
  title={Deep learning assisted measurement of echocardiographic left heart parameters: improvement in interobserver variability and workflow efficiency},
  author={Mor-Avi, Victor and Blitz, Alexandra and Schreckenberg, Marcus and Addetia, Karima and Kebed, Kalie and Scalia, Gregory and Badano, Luigi P and Kirkpatrick, James N and Gutierrez-Fajardo, Pedro and Tude Rodrigues, Ana Clara and others},
  journal={The International Journal of Cardiovascular Imaging},
  volume={39},
  number={12},
  pages={2507--2516},
  year={2023},
  publisher={Springer}
}

\end{document}